\documentclass{article}
\usepackage{arxiv}

\usepackage[utf8]{inputenc} 
\usepackage[T1]{fontenc}    
\usepackage{amsfonts}       
\usepackage{nicefrac}       
\usepackage{cellspace}
\usepackage{mathtools}
\usepackage{booktabs}
\usepackage{url}
\usepackage{caption}
\usepackage{longtable}
\usepackage{amsmath}
\usepackage{amsthm}
\usepackage{amssymb}
\usepackage{bm}
\usepackage{adjustbox}
\usepackage{wrapfig}
\usepackage{epstopdf}
\usepackage{tikz}
\usepackage{framed}
\usepackage{thmtools,thm-restate}
\usepackage{enumitem}
\usepackage{algorithmic}
\usepackage{algorithm}
\usepackage{soul}
\usepackage{tcolorbox}
\usepackage{multirow}
\usepackage{array}
\usepackage{makecell}
\usepackage[colorlinks=true,breaklinks]{hyperref}

\usepackage{mathtools}
\usepackage{bbm}
\usepackage{booktabs}       

\usepackage{float}

\usetikzlibrary{arrows}
\usetikzlibrary{shapes,backgrounds}

\title{Unsupervised collaborative learning using privileged information}

\author{Yohan Foucade\\
  Universit\'e Sorbonne Paris Nord, CNRS, Institut Galilée\\
  Laboratoire d'Informatique de Paris Nord UMR 7030, F-93430, Villetaneuse, France\\
  \url{name.surname@lipn.fr}
  \AND
  Youn{\`e}s Bennani\\
  Universit\'e Sorbonne Paris Nord, CNRS, Institut Galilée\\
  Laboratoire d'Informatique de Paris Nord UMR 7030, F-93430, Villetaneuse, France\\
  \url{name.surname@sorbonne-paris-nord.fr}}

\include{commandes}

\begin{document}
\maketitle

\begin{abstract}
In the collaborative clustering framework, the hope is that by combining several clustering solutions, each one with its own bias and imperfections, one will get a better overall solution. The goal is that each local computation, quite possibly applied to distinct data sets, benefits from the work done by the other collaborators.
This article is dedicated to collaborative clustering based on the Learning Using Privileged Information paradigm. Local algorithms weight incoming information at the level of each observation, depending on the confidence level of the classification of that observation. A comparison between our algorithm and state of the art implementations shows improvement of the collaboration process using the proposed approach.

\end{abstract}
\keywords{Unsupervised learning \and Learning from learners \and Collaborative Clustering  \and Distributed learning \and Federated learning.}


\section{Introduction}
\label{sec:introduction}
Cluster analysis is a widely used technique in data analytics. Its goal is the organization of a collection of patterns into homogenous (small intra-group variability) and distinct (large inter-group variability) clusters. There is a large variety of methods for clustering and most require all the data to be available on the same place and at the same time \cite{jain1999}. However, technical, legal, confidentiality or computational performance reasons might prevent us from gathering the data or exchanging sensitive information.

Our work is based on an advanced learning paradigm proposed by Vapnik {\em et al.} called Learning Using Privileged Information (LUPI) \cite{VapnikV09,VapnikI15a}.
In addition to examples, learners can be provided with hidden information that exists in the form of explanations, results, or comparisons, etc.
The  important  question  which  Vapnik  asks  in the supervised case is:  can  the generalization performance be improved using the privileged information? Vapnik also showed this is true in the case of SVM. 

The idea of LUPI is the following: suppose we want to establish a decision rule to determine a label based on certain characteristics $X$, but during the learning phase, in addition to $X$, we also receive additional information, referred to as "Privileged Information" $X^*$ which will not be available during the test phase.
In such a case, how can we use $X^*$ to improve learning?  In this article, we propose a new approach based on the LUPI paradigm for collaborative clustering based on knowledge sharing among a set of learners.

In collaborative clustering \cite{pedrycz2002,cornuejols2018}, we assume that we have a dataset distributed among different nodes of a network (also called sites, or experts). The goal of each expert is to obtain a clustering of its local data using findings of its remote counterparts without exchanging the data themselves. This is often done in two phases. First, a clustering algorithm is applied locally and independently on each database. This is called the local phase. Then, during the collaboration phase, data sites exchange their findings in order to try to improve their results.

The kind of information algorithms share will differ depending on what kind of data distribution one is dealing with. When objects in different data sites lie in the same feature space, then experts can share information that can be represented in the feature space itself. This task in known as vertical collaborative clustering.
In horizontal collaborative clustering, on the other hand, data sites agree on the identity of the objects but these latter are described in different feature spaces, then they can no longer share that kind of information. But they can still compare the clusters they created, using {\em e.g.} their partition matrices. A partition matrix is a $N \! \times \! C$ matrix where $N$ is the number of objects in the dataset and $C$ is the number of clusters looked for. The $k$-th row of this matrix represents the membership of each object to the $k$-th cluster.

This paper is dedicated to horizontal collaborative clustering, and our main contributions are:
\begin{itemize}
    \item An implementation of an intuitive idea for collaboration: the more doubts we have about our findings, the more weight we will give to remote findings.
    \item Collaboration with every experts at the same time. Thus there is no more need to choose which site to collaborate with at each step.
    \item Collaboration at the level of each observation. This allows algorithms to fine-tune the way they collaborate with others.
     \item Detection of examples that are difficult to cluster.
     \item Improvement of efficiency by executing in parallel the base-learning processes on subsets of the training data set. 
     \item Advantage of learning from small subsets of data that fit in main memory.
     \item Improves predictive performance by combining different learning systems each having different inductive bias.
\end{itemize}
It is organized as follows: Section \ref{sec:related_work} discusses related work. Then our main contribution --Collaborative Learning Using Privileged Information-- is introduced in Section \ref{sec:theory}. Next, we present the results obtained with different datasets. Finally, we discuss the results in Section \ref{sec:discussion}.


\section{Related Work}
\label{sec:related_work}

Research in collaborative clustering has been introduced by Pedrycz in 2002 \cite{pedrycz2002}. The author proposed a collaborative version of the Fuzzy C-Means algorithm \cite{bezdek1981}. For this purpose, the objective function of the fuzzy c-means algorithm is extended with a second term which forces a clustering based on a subset to be "aware" of other partitions.

The SAMARAH method introduced by Wemmert in \cite{wemmert_et_al_2000} can deal with heterogeneous hard clustering algorithms. This method attempts to increase similarity between clusterings through conflicts evaluation and resolution. In order to compare partitions with possibly different number of clusters, a confusion matrix is computed between each pair of algorithms. Then, a consensus is found by applying a voting algorithm to these results.

In 2015, Sublime {\em et al.} proposed an approach lifting some limits of previous works. Indeed, they introduce a framework for collaboration with heterogeneous algorithms. This method has the advantage not to require any global exchange confidence parameter, thus the impact of a distant algorithm on the local results might be different for each observation.\\
However, using hard clustering results for the construction of the confusion matrices leads to some loss of information. And every algorithm has the same weight in this process. As it makes no difference between a data site with well separated clusters and one with overlapping clusters, it allows low-performing algorithms to impede the results of the better ones.

After having proposed collaborative versions of Self Organizing Maps (SOM) and Generative Topographic Mapping (GTM), Ghassany {\em et al.} combined Fuzzy C-Means and GTM algorithms to obtain a collaborative clustering algorithm.

In \cite{SublimeMCGBC17}, Sublime {\em et al.} describe a collaboration framework for model-based clustering algorithms. They define a global likelihood function and try to optimize a proxy for this function. This process requires a weight parameter between local and external information to be fixed. 

More recently, an automated method for optimizing exchange confidence has been proposed in 2018 by Sublime {\em et al.}, and the collaboration involved 4 different views. The authors achieved successful results in detecting noisy views, but the method tends to favor collaboration between very similar views.

For a more comprehensive survey of the collaborative clustering literature, the reader may refer to \cite{cornuejols2018}.


\section{Theoretical framework and learning algorithm}
\label{sec:theory}

In this section, we describe the general framework in which our method applies and we describe our algorithm.

\subsection{Setting}

The goal of collaborative learning is to learn from local data and from a set of learners with data stored across a large number of sites.

More formally, let the global data set $X$ distributed among $P$ sites:\\
$X^{[1]}, \dots, X^{[p]}, \dots, X^{[P]}$, where $X^{[p]}=\lbrace x_i^{[p]}\rbrace_{i=1}^{N}$ is a set of $N$ objects and each object $x_i^{[p]} \in \mathbb{R}^d$ is characterized by $d$ features (i.e. is a vector in an $d$-dimensional feature space).

Our study is centered on the horizontal collaborative clustering problem. Thus, each data site has access to different $d^{[p]}$ features describing the same individuals. On each data site, there is an algorithm $\mathcal{A}^{[p]}, \, p=1, \dots, P$ trying to come up with a clustering of its data $X^{[p]}$. The process has two steps: a local step, and a collaborative step.

\subsection{Underlying framework}

\subsubsection{Local step}
During the local step, each algorithm $\mathcal{A}^{[p]}$ works on its local data set $X^{[p]}$ and fits its parameters as it would do in a non-collaborative framework. One requirement is that all algorithms have to be probabilistic ones. In particular, one of the output is a responsibility matrix $R^{[p]}$. This matrix contains the contributions of each component to each observation, $R_{i, k}^{[p]} = \mathbb{P}(Z_i = k | X_i, \theta)$, where $K$ is the number of clusters, $Z_i \in \{1,\dots, K\}$ are the components of the model and the elements of $\theta$ are the distributions parameters.

Once every model has been trained, we want them to exchange information in order to improve their performances. This is done in the collaborative step.

\subsubsection{Collaborative step}
\label{sec:collaboration}

In the collaborative steps, the different algorithms $\mathcal{A}^{[p]}$ will exchange information in order to try to improve their respective classification.
In this context, the idea of LUPI is the following: suppose we want to establish a clustering based on certain characteristics $X^{[p]}$, but during the learning phase, in addition to $X^{[p]}$, we also receive additional information, $X^{{[p]}^*}$. This additional information is presented in the form of partition matrices $R^{[p]}$. The local learner therefore has access to $X^{[p]}$ and $X^{{[p]}^*}$, instead of only $X^{[p]}$, to do its clustering.
In our case, for each algorithm $\mathcal{A}^{[p]}$, $X^{{[p]}^*} \equiv R^{-[p]}$, where:
\begin{equation}
    R^{-[p]} = \{R^{[q]} \, : \, q \in \{1, \dots, P\} \setminus \{p\}\}
\end{equation}

The set $R^{-[p]}$ contains all partition matrices of site $p$'s counterparts.
In this case, how can we use $X^{[p]}$ and $X^{{[p]}^*}$ to improve each local learners $\mathcal{A}^{[p]}$?

\bigskip

For the sake of simplicity, we consider two learners (sites): $P=2$. On the local step, they yield two partition matrices: $R^{[1]}(t)$ and $R^{[2]}(t)$. In particular, for sample $x_i$, we have $R^{[1]}_{i,.}$ and $R^{[2]}_{i,.}$.
Thus, the update rule for learner 1 for sample $x_i$ is:
\begin{equation}\label{eq:general_form}
    R^{[1]}_{i,.}(t+1) \longleftarrow f\left(R^{[1]}_{i,.}(t), R^{[2]}_{i,.}(t)\right)
\end{equation}

We want this update rule to depend on the level of certainty data site 1 has about its opinion. The higher this level, the lower the difference between $R^{[1]}_{i,.}(t+1)$ and $R^{[1]}_{i,.}(t)$. Conversely, the lower it is, the more weight will be put on $R^{[2]}_{i,.}(t)$. Moreover, the higher (resp. the lower) site $2$'s certainty, the more (resp. less) it will influence $R^{[1]}_{i,.}(t+1)$.

It turns out that the probabilistic framework is endowed with a measure that can be interpreted as the amount of uncertainty in a distribution, namely the entropy. It is defined as follows:

\begin{equation}
    H(X) = -\sum_{i=1}^{d}p(x_i)\log_2p(x_i)
\end{equation}

Where $X$ is a random variable with possible values $\{x_1, \dots, x_d\}$.
It follows that the uncertainty for a distribution $R_i$ is:
\begin{equation}
    H(R_{i,.}) = - \sum_{k=1}^{K} R_{i, k} \log_2 R_{i, k}
\end{equation}

And its normalized version, using the fact that it is positive and maximized when $X$ is uniformly distributed:
\begin{equation}
    \mathcal{H}(R_{i,.}) = \frac{H(R_{i,.})}{log_2(K)}
\end{equation}

Then $0 \leq \mathcal{H}(R_{i,.}) \leq 1$, and
\begin{itemize}
    \item $\mathcal{H}(R_{i,.}) \approx 0 \Rightarrow R_{i,.} \approx \mathbbm{1}_{Z_i=k} \textrm{ for some } k \in {1,\dots,K}$, i.e. we have high confidence about membership of observation $i$.
    \item $\mathcal{H}(R_{i,.}) \approx 1 \Rightarrow R_{i,.} \approx \frac{1}{K} \ \forall k \in {1,\dots,K}$, i.e. we have high uncertainty about membership of observation $i$.
\end{itemize}

Equation \ref{eq:collab_eq} gives the update rule for $R^{[ii]}$ when $P$ algorithms are involved in the collaboration.

\begin{equation}\label{eq:collab_eq}
    R^{[p]}(t+1)\!\longleftarrow \alpha^{[p]} \cdot \! R^{[p]}(t) + \sum\limits_{R^{[q]}(t) \in R^{-[p]}} \beta^{[p]}_{[q]} \cdot R^{[q]}(t)
\end{equation}

where

\begin{equation}
     \left\{
     \begin{array}{lll}
        \alpha^{[p]} \!=\! \left(\frac{1}{P-1} \! \sum\limits_{{R^{[q]}(t) \in R^{-[p]}}} \! \mathcal{H}\! \left( \!R^{[q]}(\!t\!)\!\right)\right) \!\cdot\! \left(\!1\!-\!\mathcal{H}\!\left(\!R^{[p]}(\!t\!)\!\right)\right)
\\
\\
    \beta^{[p]}_{[q]} = \mathcal{H}\left(R^{[p]}(t)\right)\!\cdot\!\left(1-\mathcal{H}\left(R^{[q]}(t)\right)\right)
     \end{array}
     \right.
\end{equation}

In these equations, $\alpha^{[p]}$ is a vector of size $N$. Each element $i$ of $\alpha^{[p]}$ is a weight associated with the local classification of the $i$-th observation. This weight depends negatively on the amount of uncertainty carried by the local classification, and positively on the average remote classification uncertainty.

Similarly, $\beta^{[p]}_{[q]}$ are vectors of size $N$. Each element $i$ of $\beta^{[p]}_{[q]}$ is a weight associated with the classification of the $i$-th observation by algorithm $\mathcal{A}^{[q]}$. This weight depends positively on the amount of uncertainty carried by the local classification, and negatively on the uncertainty of the remote classification.

Note that the data sites only share there partition matrices. Therefore, no matter what the underlying algorithms are, provided that they look for the same number of components, the collaboration algorithm is still relevant. However, the data sites need to agree on the identity of the clusters. Our implementation uses the Hungarian algorithm to reorder the clusters on each data site. Next, we show how the values of $\alpha$ and $\beta$ can be used to visualize information flow in the collaboration process, and describe the CoLUPI algorithm.

\subsubsection{Collaborative learning algorithm}

Based on the theoretical formalism developed and presented in the previous section, we can design a learning algorithm to establish exchanges between the different sites through a collaborative process. This algorithm therefore uses the equation \ref{eq:collab_eq} for updating the parameters of the sites in collaborative interaction.

\begin{algorithm}\label{algorithm}
\caption{CoLUPI algorithm}
\begin{algorithmic}
    \STATE \textbf{Local step:}\\
        \FORALL{algorithms $\mathcal{A}^{[p]}$}
            \STATE train $\mathcal{A}^{[p]}$ on the data $X^{[p]}$
            \STATE get local parameters $\theta^{[p]}$ and partition matrix $R^{[p]}$
        \ENDFOR
    \STATE \textbf{Collaborative step:}\\
        \REPEAT
            \FORALL{algorithms $\mathcal{A}^{[p]}$}
                \STATE{$R^{[p]}(t+1) \gets f\left(R^{[p]}(t), R^{-[p]}(t)\right)$}
                \STATE train model starting from new partition $R^{[p]}(t+1)$ using data $X^{[p]}$
                \STATE get $\theta^{[p]}(t\!+\!1)$ and $R^{[p]}(t\!+\!1)$
                \IF{quality$\left(R^{[p]}(t\!+\!1)\right)$ is better than quality$\left(R^{[p]}(t)\right)$}
                    \STATE Accept collaboration
                \ELSE
                     \STATE $\theta^{[p]}(t\!+\!1) \gets \theta^{[p]}(t)$
                     \STATE $R^{[p]}(t\!+\!1) \gets R^{[p]}(t)$
                \ENDIF
                
            \ENDFOR
        \UNTIL{no algorithm improves for its criterion}
\end{algorithmic}
\end{algorithm}

\subsection{Visualization of the collaboration process}

\paragraph{}
In this section, we are interested in visualizing the information flow during collaboration. As it has been mentioned above, each collaborator assigns one weight to each source data site (including itself), for each observation. The mean of these weights for each remote site gives us the average weight assigned to that site and can be regarded as a confidence coefficient, except that it is not uniform across observations. If there are $P$ sites, then this gives us a $P\!\times\!P$ matrix as defined in \ref{eq:collab_matrix} which can be regarded as a confidence matrix. We show this collaboration heatmap for the Wdbc dataset in Section \ref{sec:results}.

\begin{equation}
\label{eq:collab_matrix}
C_{(P \times P)} = 
\begin{pmatrix}
\overline{\alpha^{[1]}} & \overline{\beta^{[1]}_{[2]}} & \cdots & \overline{\beta^{[1]}_{[P]}} \\
\\
\overline{\beta^{[2]}_{[1]}} & \overline{\alpha^{[2]}} & \cdots & \overline{\beta^{[2]}_{[P]}} \\
\vdots  & \vdots  & \ddots & \vdots  \\
\overline{\beta^{[P]}_{[1]}} & \overline{\beta^{[P]}_{[2]}} & \cdots & \overline{\alpha^{[P]}} 
\end{pmatrix}
\end{equation}
Where 
\begin{equation}
    \overline{\alpha^{[p]}} = \frac{1}{N} \sum_{i=1}^{N}\alpha_i^{[p]}\text{,}
\end{equation}
and
\begin{equation}
    \overline{\beta^{[p]}_{[q]}} = \frac{1}{N} \sum_{i=1}^{N}{\beta_i}^{[p]}_{[q]}.
\end{equation}


\section{Experimental validation}
\label{sec:results}

In this section, we present the results obtained after running our algorithm on several datasets. First, we describe the datasets that has been used in the experiments, then we present two kinds of results: performance assessment and visualization of the collaboration process. Finally, we discuss the execution time of our algorithm with different input values.

\subsection{Data sets}
\label{sec:datasets}

\begin{itemize}
    \item The Breast Cancer Wisconsin (Diagnostic) (WDBC) dataset consists in 569 digitized images of a breast mass. There are 30 real-valued input variables describing the cell nuclei present in each image. Each observation is labelled as benign or malignant.
    %
    \item The Spambase data set consists in 57 attributes describing a collection of 4601 spam and non-spam emails.
    %
    \item The Battalia3 data set is an artificial data set describing 2000 generated exoplanets with 27 numerical attributes.
    %
    \item The MV2 dataset features 2000 data points, each described by 6 features. They have been randomly generated from a mixture of one noise and four Gaussian components.
    %
    \item The Isolated letters (Isolet) data set has 617 input variables describing 7797 voice recordings of individuals who spoke the name of each letter of the alphabet.
    %
    \item The Madelon data set is an artificial data set containing 4400 data points grouped in 32 clusters placed on the vertices of a five dimensional hypercube. Then 15 redundant features and 480 useless features (random probes) were added, for a total of 500 attributes.
\end{itemize}

\subsection{Experimental protocol}
In our experiments, we split the databases in order to obtain a horizontal collaborative clustering setting -- i.e. each data site has access to different input variables for the same set of observations.


\subsection{Results and analysis}
\label{sec:subresults}

\subsubsection{Co-LUPI using GTM models}
\begin{figure*}
    \centering
    \includegraphics[width=.9\textwidth]{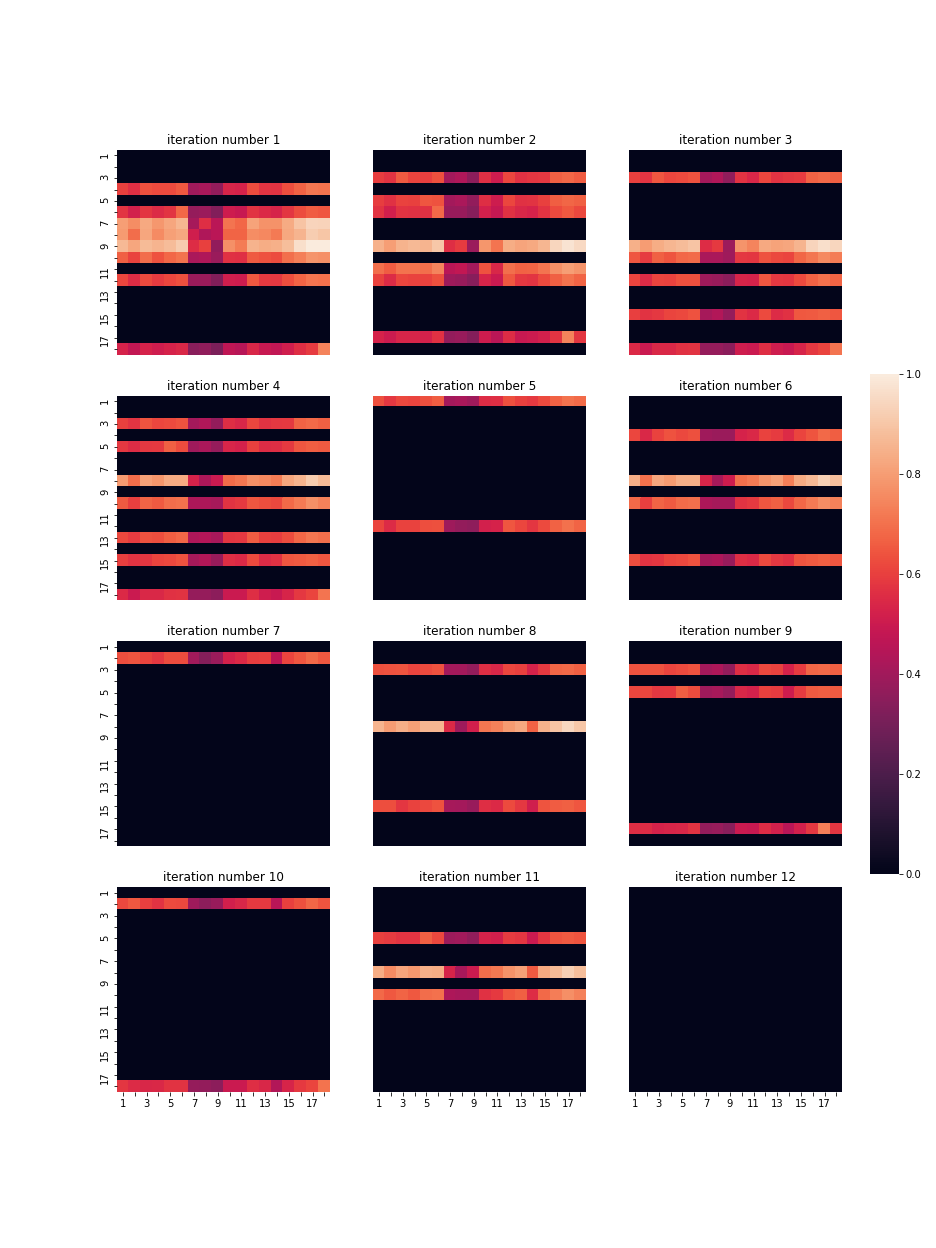}
    \caption{Information flow during collaboration for the Wdbc data set. Each row represents a data site and each column the total weight assigned to that source.}
    \label{fig:collab_heatmap}
\end{figure*}

The Co-LUPI algorithm has been applied to each of the 6 data sets mentioned in \ref{sec:datasets} using Generative topographic mappings.
The criterion for acceptance of collaboration was the improvement of the Davies-Bouldin index. This is an internal index, thus it does not require prior knowledge on the data structure. A second version of the algorithm, RCo-LUPI, has been implemented. It features a new, random, initialization of the responsibility matrix at each step, along with the collaboration matrix. This technique is often used in unsupervised learning, it is meant to reduce dependence on the initial parameters. Table \ref{tab:results_comparison} shows that in most cases, RCo-LUPI performed slightly better than Co-LUPI.\\
In order to visualize the dynamic of this process, one can look at the successive confidence matrices of the collaboration step. Figure \ref{fig:collab_heatmap} represents such data. The Co-LUPI algorithm has been applied to the WDBC data set, split among 18 data sites. Obviously, not all algorithms benefited from the collaboration at each step. In particular, the algorithm running on data site number one did not improve its results until the fifth iteration in the collaboration step. Furthermore, on iteration number 7, only the second algorithm did improve. While this can be interpreted as a sign of an imminent end of the
\begin{table*}
\center
\caption{Experimental results, Davies-Bouldin index - Co-EM: \cite{SublimeMCGBC17}, Co-MV: \cite{GhassanyGB13}, Co-GTM: \cite{GhassanyGB12b}, Co-SOM: \cite{GrozavuB10}, Co-LUPI, RCo-LUPI}\label{tab:results_comparison}
\begin{tabular}{ccccccc}
    \hline
    Name & \rotatebox{45}{Co-EM} & \rotatebox{45}{Co-MV}  & \rotatebox{45}{Co-GTM} & \rotatebox{45}{Co-SOM} & \rotatebox{45}{Co-LUPI} & \rotatebox{45}{RCo-LUPI}\\
    \hline
        &&&&&\\
    \begin{tabular}{l}Wdbc\end{tabular}         & 0.85 & 0.97 & 0.9  & 0.84 & 0.78          & \textbf{0.69}\\
        &&&&&\\
    \begin{tabular}{l}Spam Base\end{tabular}    & 0.94 & 1.27 & 0.92 & 0.87 & \textbf{0.42} & 0.59\\
        &&&&&\\
    \begin{tabular}{l}Battalia3\end{tabular}    & 2.43 & 2.83 & 2.68 & 2.51 & 1.47          & \textbf{1.37}\\ 
        &&&&&\\
    \begin{tabular}{l}MV2\end{tabular}          & 1.34 & 1.34 & 1.61 & 1.44 & 0.86          & \textbf{0.85}\\
        &&&&&\\
    \begin{tabular}{l}Isolet\end{tabular}       & --   & --   & --   & --   & 1.33          & \textbf{1.31}\\
        &&&&&\\
    \begin{tabular}{l}Madelon\end{tabular}      & --   & --   & --   & --   & 0.87          & \textbf{0.82}\\
    \bottomrule
\end{tabular}
\end{table*}
process, 3 other algorithms benefited from these new findings on the next iteration. The process did not end before 4 more iterations.

\subsection{Comparison with other collaborative approaches}
The Co-LUPI and RCo-LUPI algorithms were empirically compared to four recent implementations of collaborative clustering algorithms.
The optimization process behind the Co-EM algorithm is based on variational EM. It optimizes a collaborative term which is equivalent to the entropy \cite{SublimeMCGBC17}. The same principle is used in Co-EM with the difference that it is based on prototypes, while the Co-EM is based on partitions \cite{GhassanyGB13}. In the Co-SOM and Co-GTM methods, the SOM and GTM loss functions where modified in order to penalize the difference between local and remote parameters \cite{GrozavuB10,GhassanyGB12b}.

\begin{figure}
    \caption{Friedman and Nemenyi test for comparing multiple approaches over multiple data sets: Approaches are ordered from left (the best) to right (the worst)}
    \label{fig:FriedmanTest}
    \centering
    \includegraphics[scale = 0.3463]{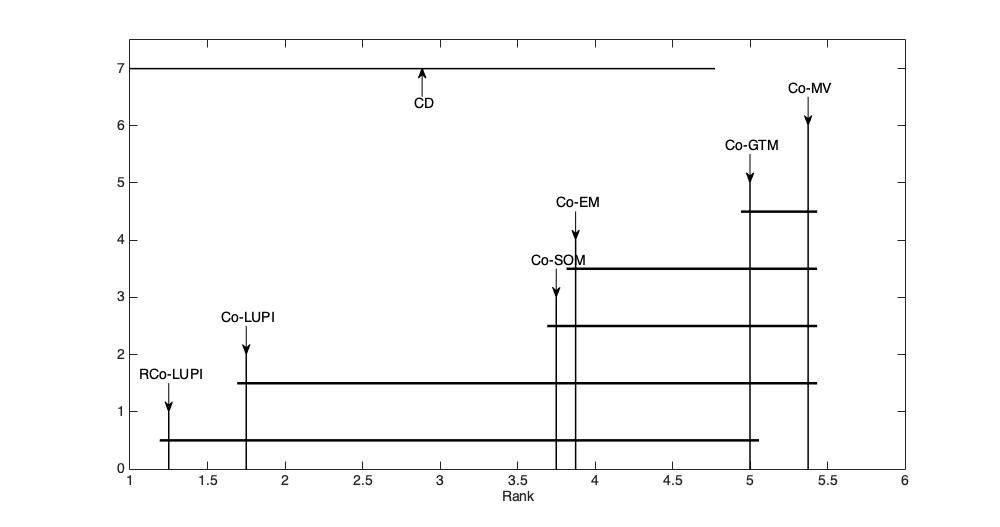}
   
\end{figure}

In order to assess the performance of our approaches, we use the Friedman test and Nemenyi test recommended in \cite{Demsar06}. First, algorithms are ranked according to their performance on each dataset. There are as many rankings as their are datasets. Then, the Friedman test is conducted to test the null-hypothesis under which all approaches are equivalent, and in this case their average ranks should be equal. If the null hypothesis is rejected, then the Nemenyi test will be performed. If the average ranks of two approaches differ by at least the critical difference(CD), then it can be concluded that their performances are significantly different. In the Friedman test, we set the significance level $ \alpha=0.05$. The figure \ref{fig:FriedmanTest} shows a critical diagram representing a projection of average ranks of the algorithms on enumerated axis. The methods are ordered from left (the best) to right (the worst) and a thick line connects the groups of algorithms that are not significantly different (for the significance level  $ \alpha=5\%$). As shown in figure \ref{fig:FriedmanTest}, Co-LUPI and RCo-LUPI seem to achieve some improvement over the other proposed techniques. But the results are not sufficient to conclude to a statistically significant improvement. This result can be explained by the small number of datasets and by the fact that the Nemenyi test only considers algorithms performances through their ranks and is blind to the actual value of the performance index.

\section{Conclusion and future work}
\label{sec:discussion}
We introduced Co-LUPI and RCo-LUPI, algorithms based on usage of privileged information and probability for collaborative clustering. This allows local algorithms to fine-tune collaboration depending on their (un)certainty --and their remote counterparts' (un)certainty-- about the classification of each data point, as measured by the entropy. This update rule is straightforward and the collaboration is done with every remote site at the same time. This avoids the classical problem of choosing which site to collaborate with at each step.\\
We tested our approach on several data sets in the horizontal collaboration setting, but it is also applicable in the hybrid setting. The results showed improvement over state of the art. The framework also provides a way to visualize information flow during the process. It exhibited interesting behaviors, as algorithms with lower initial performance tended to make heavier use of incoming information than the other ones. Moreover, even the algorithms with the best results after the local step were able to improve in the process. Indeed, the flexibility brought by Co-LUPI in the weighting of incoming information allows algorithms to benefit from globally less efficient counterparts, because the latter can be locally more efficient.\\
These results are very promising and the algorithm can be improved in several ways. One big improvement would be to relax the assumption that every algorithm look for the same number of clusters. This might be challenging as the algorithms also have to agree on the identity of the different clusters. This problem is related to the generalized assignment problem.
We will also investigate how is the CoLUPI algorithm performing when used with some of the most representative algorithms to solve the problem, e.g. monarch butterfly optimization (MBO) algorithm \cite{monarch_butterfly}.


\end{document}